\documentclass{article}

\usepackage[preprint]{neurips_2024}

\usepackage[utf8]{inputenc} 
\usepackage[T1]{fontenc}    
\usepackage{hyperref}       
\usepackage{url}            
\usepackage{booktabs}       
\usepackage{amsfonts}       
\usepackage{nicefrac}       
\usepackage{microtype}      
\usepackage{xcolor}         

\usepackage{multirow}
\usepackage{graphicx}
\usepackage{amsmath}
\usepackage{tabularx}

\title{Empathy-R1: A Chain-of-Empathy and Reinforcement Learning Framework for Long-Form Mental Health Support}

\author{%
  Xianrong~Yao\thanks{Equal contribution.} \\
  School of Future Technology\\
  South China University of Technology\\
  \And
  Dong~She\footnotemark[1] \\
  School of Future Technology\\
  South China University of Technology\\
  \And
  Chenxu~Zhang \\
  School of Future Technology\\
  South China University of Technology\\
  \And
  Yimeng~Zhang \\
  School of Future Technology\\
  South China University of Technology\\
  \And
  Yueru~Sun \\
  School of Future Technology\\
  South China University of Technology\\
  \And
  Noman~Ahmed \\  
  School of Future Technology\\
  South China University of Technology\\
  \And
  Yang~Gao \\
  School of Future Technology\\
  South China University of Technology\\
  \And
  Zhanpeng~Jin\thanks{Corresponding author.} \\
  School of Future Technology\\
  South China University of Technology\\
}

\begin{document}

\maketitle

\begin{abstract}
Empathy is critical for effective mental health support, especially when addressing Long Counseling Texts~(LCTs). However, existing Large Language Models~(LLMs) often generate replies that are semantically fluent but lack the structured reasoning necessary for genuine psychological support, particularly in a Chinese context. To bridge this gap, we introduce \textbf{Empathy-R1}, a novel framework that integrates a Chain-of-Empathy (CoE) reasoning process with Reinforcement Learning~(RL) to enhance response quality for LCTs. Inspired by cognitive-behavioral therapy, our CoE paradigm guides the model to sequentially reason about a help-seeker's emotions, causes, and intentions, making its thinking process both transparent and interpretable. Our framework is empowered by a new large-scale Chinese dataset, \textbf{Empathy-QA}, and a two-stage training process. First, Supervised Fine-Tuning instills the CoE's reasoning structure. Subsequently, RL, guided by a dedicated reward model, refines the therapeutic relevance and contextual appropriateness of the final responses. Experiments show that Empathy-R1 achieves strong performance on key automatic metrics. More importantly, human evaluations confirm its superiority, showing a clear preference over strong baselines and achieving a Win@1 rate of \textbf{44.30\%} on our new benchmark. By enabling interpretable and contextually nuanced responses, Empathy-R1 represents a significant advancement in developing responsible and genuinely beneficial AI for mental health support.
\end{abstract}

\section{Introduction}

Mental health is a pressing global issue. Conditions like depression are a leading cause of disability, affecting 5\% of adults worldwide and tens of millions in China alone, creating a staggering demand for support amidst a significant "treatment gap" due to stigma and resource scarcity \citet{who_depression, who2023mentalhealth}. Consequently, online platforms have become a critical first-stop for users to articulate their struggles in accessible, anonymous spaces \cite{white2001receiving, eysenbach2004health}.

This online help-seeking behavior often manifests as what we term \textbf{Long Counseling Texts (LCTs)}: detailed, lengthy posts that are far more than simple queries. Unlike short empathic conversations, LCTs present vivid self-narratives of complex personal histories and profound emotional distress. Responding effectively to LCTs is a significant challenge; generic comforting phrases are insufficient and can feel invalidating. The task instead demands a deep, analytical approach to deconstruct the user's situation, identify underlying cognitive patterns, and offer structured, counselor-like guidance.

With the rise of Large Language Models (LLMs), a promising avenue has opened for automating mental health support \cite{xiao2024healme, almakinah2025enhancing, na2024cbt}. However, applying existing LLMs to the complex domain of LCTs reveals fundamental challenges. First, they often struggle to translate semantic fluency into genuine therapeutic depth. While capable of generating seemingly empathetic text, their responses can feel generic, failing to consistently integrate professional knowledge with stable empathy. Second, their therapeutic strategy is often implicit and unstable. Lacking an explicit, psychologically-grounded plan, most models generate replies in an opportunistic manner, which severely limits the therapeutic value of the interaction. This predicament is exacerbated by a scarcity of high-quality, large-scale public datasets for LCTs, especially in Chinese. Existing resources like PsyQA \cite{sun2021psyqa}, while valuable, are constrained by a limited scope (covering only 9 main topics) and do not fully reflect contemporary issues.

To bridge this critical gap, we propose Empathy-R1, a framework that advances the concept of a Chain-of-Empathy (CoE), a reasoning paradigm previously explored to structure empathetic responses \cite{lee2023chain}. While this prior work established a valuable direction, its high-level framework has not been specifically operationalized for the unique complexities of LCTs. Our work develops CoE into a more granular and psychologically-grounded paradigm. Inspired by clinical techniques from Cognitive Behavioral Therapy (CBT), and distinct from general-purpose methods like Chain-of-Thought (CoT), our CoE operationalizes a counselor’s thinking process through a detailed four-layered analysis: (L1) Emotions and Context, (L2) Causes and Beliefs, (L3) Intent Analysis, and (L4) Response Strategy. This specific structure makes the model’s reasoning transparent, interpretable, and directly aligned with therapeutic practices. To power this advanced framework, we also construct and release Empathy-QA, a new large-scale Chinese dataset of LCTs tailored for this complex task.

Our contributions can be summarized as follows:
\begin{enumerate}
    \item We propose Empathy-R1, a framework centered on our novel Chain-of-Empathy (CoE) paradigm. This framework teaches LLMs the multi-layered reasoning of a human counselor to provide deeply reasoned, rather than superficial, support.
    \item We construct and release Empathy-QA, a large-scale, contemporary Chinese dataset for LCTs, addressing a critical resource gap for developing and evaluating mental health support models.
    \item Through rigorous human evaluations, we validate that Empathy-R1 is overwhelmingly preferred over strong baselines. Our model establishes a new state-of-the-art, achieving a 44.30\% Win@1 rate and proving its effectiveness in generating genuinely helpful support.
\end{enumerate}

\section{Empathy-QA: Mental Health Support Corpus}

In the field of single-turn long-form Chinese mental health counseling response corpora, PsyQA \cite{sun-etal-2021-psyqa} represents an important early Chinese dataset. However, as it was collected several years ago, it does not fully capture the evolving expressions of psychological distress among contemporary Chinese users. To address this gap, we introduce \textbf{Empathy-QA}, a new large-scale dataset crawled from a curated list of public Chinese professional counseling platforms and social media forums. \textbf{Further details on the data collection process are provided in the appendix.}

To highlight its timeliness, Empathy-QA includes contemporary issues not present in PsyQA. For instance, it includes some discussions related to psychological distress stemming from the rapid development of large models, such as a user asking, \textit{“I work in video editing and special effects, and I'm very anxious about the rise of AI video. I'm worried that my skills will become obsolete and that I'll be replaced. This fear is even making me doubt my career choice. How should I deal with this fear and find a new direction?”}—a theme absent in earlier corpora.

The dataset includes 40,959 user questions and 168,470 long-form responses. To ensure data quality and ethical standards, we implemented a rigorous filtering and anonymization process. We removed noisy content, advertisements, and non-psychological inquiries, retaining only samples with clearly expressed psychological issues and responses exceeding 100 characters (see Table~\ref{tab:dataset-stats} for detailed statistics).

Each data instance in \textbf{Empathy-QA} consists of four main fields: \textit{question title}, \textit{question description}, \textit{topic label}, and one or more \textit{response texts}. The \textit{title} and \textit{description} represent the user's query, the \textit{topic label} categorizes the issue, and each \textit{response} field contains content from a human respondent.

\begin{table}[thb]
\centering
\setlength{\tabcolsep}{1.5mm}
\caption{Comparison of dataset statistics between PsyQA and Empathy-QA.}
\begin{tabular}{lcc}
\hline
\textbf{Criteria} & \textbf{PsyQA} & \textbf{Empathy-QA} \\
\hline
\# Questions & 22,346 & 40,959 \\
\# Answers & 56,063 & 168,470 \\
\# Main topics & 9 & 29 \\
\# Subtopics & – & 182 \\
Avg. chars per question & 21.6 & 22.7 \\
Avg. chars per description & 168.9 & 275.9 \\
Avg. chars per answer & 524.6 & 775.6 \\
Avg. responses per question & 2.51 & 4.11 \\
\hline
\end{tabular}
\label{tab:dataset-stats}
\end{table}

\section{Empathy-R1}

\begin{figure}
    \centering
    \includegraphics[width=1\linewidth]{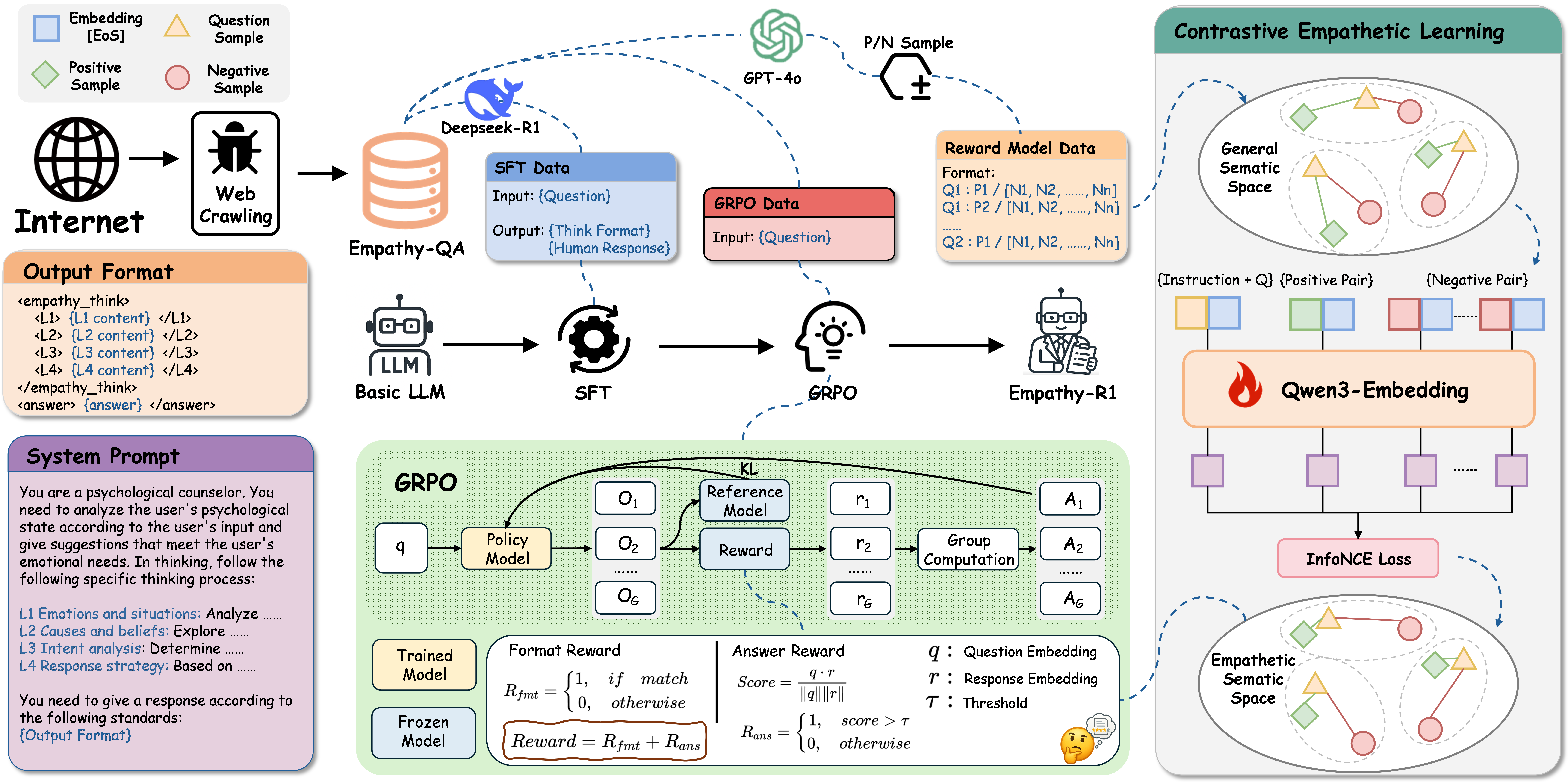}
    \caption{Overview of the Empathy-R1 construction pipeline. A base model first undergoes Supervised Fine-Tuning (SFT) to adopt the structured Chain-of-Empathy (CoE) format. Subsequently, Group Relative Policy Optimization (GRPO) refines the model's therapeutic quality, guided by a reward model. This reward model is trained using Contrastive Empathetic Learning to score the empathetic relevance of responses.}
    \label{fig:framework}
\end{figure}

\subsection{Chain-of-Empathy}

The Chain of Empathy (CoE), the core of our Empathy-R1 framework, is a psychologically-grounded reasoning paradigm designed to integrate established psychotherapeutic principles into the reasoning phase of large language models (LLMs). The primary motivation is to equip the model with a structured, multi-layered approach to understanding user emotions—mirroring a human therapist's methodical process—and thus move beyond superficial empathy.

To design the CoE, we reviewed literature from foundational psychological therapies and identified a common procedural pattern for understanding a client. Based on this, we developed a four-tiered progressive analysis framework. The first tier, \textbf{L1: Emotions and Context}, requires the model to identify the user's core emotions within their objective situation. This is grounded in Appraisal Theory~\cite{smith1990emotion}, which posits that emotions arise from an individual's subjective evaluation of an event, making the link between emotion and context fundamental for accurate understanding. Next, \textbf{L2: Causes and Beliefs}, delves deeper to explore the underlying reasons and potential cognitive biases. This stage is inspired by the concept of `mentalizing` or Empathic Accuracy~\cite{ickes1997studying}, which involves not just recognizing an emotion but correctly inferring the thoughts and beliefs causing it. Subsequently, \textbf{L3: Intent Analysis}, aims to discern the user's primary communication goal. Drawing from Davis's multi-dimensional view of empathy~\cite{davis1983measuring}, this step recognizes that users may seek different forms of support, such as simple validation, shared understanding, or actionable advice. Finally, \textbf{L4: Response Strategy}, synthesizes insights from the previous layers to formulate a therapeutic response. This is informed by established clinical communication frameworks like Carl Rogers's Active Listening principles~\cite{rogers1957necessary} and the NURSE model~\cite{back2009mastering}, ensuring the generated response is strategically aligned with the user's needs.

In our engineering implementation, we use a customized system prompt (see Figure~\ref{fig:framework}) to guide the model's thought process. The prompt explicitly instructs the model to follow the four-layered CoE thinking process (L1: Emotion and Context, L2: Cause and Belief, L3: Intent Analysis, L4: Response Strategy). This structured "thinking" process is encapsulated within \texttt{<empathy\_think>} tags, followed by the final \texttt{<answer>} tag containing the user-facing response. This method renders the model's empathetic reasoning process both transparent and interpretable.

In summary, the CoE framework provides a deliberate and theoretically-grounded pathway for LLM reasoning. This design moves the model beyond superficial pattern matching, compelling it to follow a structured process that mirrors established therapeutic practices, thus laying a robust foundation for generating higher-quality empathetic responses.

\subsection{Supervised Fine-Tuning}

To construct the necessary training data, we developed a targeted generation and review process. First, we randomly selected question-answer pairs from the Empathy-QA training set. For each pair, we prompted the Deepseek-R1 model to generate a corresponding CoE thinking process, effectively creating a structured rationale that connects the user's query to the existing high-quality response. A psychology professional then meticulously reviewed these generated chains for quality and therapeutic validity. This process yielded a curated dataset of 200 high-quality instances, each containing a question, an human answer, and a professionally-vetted CoE chain.

This initial stage functions as a form of behavioral cloning, compelling the model to internalize the structural scaffolding of professional counseling. As this phase is a cold-start intended to teach the model the CoE architecture and format, a smaller, high-quality dataset is sufficient. We implement this using a single-turn conversational fine-tuning format. Each training instance begins with a detailed \texttt{system} prompt that defines the model's persona, its task, and the explicit four-layered CoE reasoning structure (L1-L4). This is followed by the \texttt{user}'s help-seeking query. The model is then trained to generate the \texttt{assistant}'s response, which is a composite of the generated \texttt{<empathy\_think>} chain and the original human-written \texttt{<answer>}.

The strategic goal of this phase is not to achieve optimal performance, but rather to create a structurally-aligned agent. While this imitation learning is crucial for instilling the desired reasoning format, it can, as our later ablation studies suggest, lead to outputs that are somewhat rigid. Therefore, this SFT stage is fundamental as it creates a predictable and structurally sound agent, providing a high-quality starting point for the subsequent Reinforcement Learning phase, which is then tasked with refining therapeutic nuance and empathetic quality.

\subsection{Reinforcement Learning with CoE}

In the Reinforcement Learning (RL) phase, we refine the SFT model—based on Qwen3-8B~\cite{qwen3}—to improve empathetic reasoning. Since mental health support often allows multiple valid responses, we adopt Group Relative Policy Optimization (GRPO)~\cite{guo2025deepseek}, which optimizes the model based on the relative quality of multiple responses generated per query. As shown in Figure~\ref{fig:framework}, our \textbf{Empathy-R1} framework evaluates each output using a composite reward that captures both adherence to the Chain-of-Empathy (CoE) structure and the quality of the final answer, with updates guided by group-relative advantages (Eq.~\eqref{eq:relative-adv}). This setup encourages contextual reasoning before response generation. We next describe the GRPO algorithm and our reward design in detail.

\subsubsection{Group Relative Policy Optimization}
\label{sec:grpo}

To promote the generation of emotionally aligned reasoning chains, we employ \textit{Group Relative Policy Optimization} (GRPO). Unlike methods that evaluate individual rewards independently, GRPO considers the relative quality of sampled outputs within a batch.

At each training step, the current policy $\pi_{\text{old}}$ generates $G$ candidate outputs $\{o_i\}_{i=1}^G$. Each output $o_i$ comprises the full text generated by the model, including both the structured thinking process and the final answer. It is assigned a reward $r_i$ that reflects both format correctness and answer quality (see details in Sec.~\ref{sec:reward_design}). The group-relative advantage $A_i$ is computed as:
\begin{equation}
A_i = \frac{r_i - \mu_{\{r_1, r_2, \cdots, r_G\}}}{\sigma_{\{r_1, r_2, \cdots, r_G\}}}
\label{eq:relative-adv}
\end{equation}
where $\mu_{\{r_1, r_2, \cdots, r_G\}}$ and $\sigma_{\{r_1, r_2, \cdots, r_G\}}$ denote the mean and standard deviation of rewards within the group. This normalization emphasizes outputs that are not only high-quality in absolute terms but also superior relative to other candidates in the same batch.

Outputs with above-average rewards receive higher advantage values, encouraging the policy to optimize toward relatively better responses. The policy is then updated by maximizing the following objective:

\begin{align}
\mathcal{J}_{\text{GRPO}}(\theta) =\; & 
\mathbb{E}_{q \sim P(Q), \{o_i\}_{i=1}^G \sim \pi_{\theta_{\text{old}}}(O|q)} \Bigg[ \frac{1}{G} \sum_{i=1}^{G} \Bigg( \nonumber \\
& \min \left( 
r_i^{ratio} A_i,\,
\text{clip} \left(
r_i^{ratio},\, 
1 - \epsilon,\, 1 + \epsilon 
\right) A_i 
\right) \Bigg) \nonumber \\
& \quad - \beta \mathbb{D}_{\text{KL}}\left( \pi_\theta \,\|\, \pi_{\text{ref}} \right)
\Bigg]
\end{align}

Where $r_i^{\text{ratio}} = \frac{\pi_\theta(o_i \mid q)}{\pi_{\theta_{\text{old}}}(o_i \mid q)}$ is the probability ratio between the new and old policies, encouraging higher-quality outputs when $A_i$ is large.
The clipping operator $\text{clip}(r_i^{\text{ratio}},\, 1 - \epsilon,\, 1 + \epsilon)$ stabilizes training by preventing excessively large policy updates.
The KL divergence term $\mathbb{D}_{\text{KL}}(\pi_\theta \,\|\, \pi_{\text{ref}})$ constrains the new policy from deviating too far from the reference policy, which is the model from the SFT phase.
The coefficient $\beta$ is a hyperparameter controlling the strength of the KL penalty.

\subsubsection{Reward Design}
\label{sec:reward_design}

Our composite reward function is designed to align the model's outputs with two distinct but complementary goals: adhering to the structured Chain-of-Empathy (CoE) format and producing a high-quality, empathetic final answer. The total reward is computed as:
\begin{equation}
r_{\text{total}} = r_{\text{format}} + r_{\text{answer}}
\end{equation}
where both $r_{\text{format}}$ and $r_{\text{answer}}$ are normalized to the $[0, 1]$ range. The reinforcement learning process optimizes the entire generated output, but the reward components target specific aspects of the generation.

\paragraph{Format Reward.}
The $r_{\text{format}}$ component ensures that the model learns to follow the CoE reasoning structure. This reward is calculated based on the entire generated output string. We perform a regular expression match to verify the presence and correctness of the overall structure, specifically checking for the \texttt{<empathy\_think>} block containing all four sub-tags (\texttt{<L1>} to \texttt{<L4>}) and the final \texttt{<answer>} block. This is a binary reward:
\begin{equation}
r_{\text{format}} = 
\begin{cases} 
1, & \text{if the output format is correct} \\
0, & \text{otherwise}
\end{cases}
\end{equation}
This component explicitly guides the model to produce transparent and interpretable reasoning chains before generating the user-facing response.

\paragraph{Answer Reward}

The $r_{\text{answer}}$ component evaluates the empathetic quality of the final response, focusing exclusively on the content within the \texttt{<answer>} tags. To achieve this, we first train a dedicated reward model using a contrastive learning objective. The goal is to train a model that can distinguish between high-quality (positive) and low-quality (negative) responses for a given user query.

For each question $q$, the high-quality response from a professional counselor serves as the positive sample ($a^+$). We construct a set of diverse negative samples ($a^-$) by including: (1) responses with mismatched emotions, (2) generic, high-frequency sentences from the corpus, and (3) other responses from within the same training batch. The reward model is then trained with a margin-based triplet loss:

\begin{align}
\mathcal{L}_{\text{contrastive}} = \max(&0, \cos(E(q), E(a^-)) \nonumber \\ 
& - \cos(E(q), E(a^+)) + m)
\label{eq:contrastive_loss}
\end{align}

where $E(\cdot)$ is the embedding function of our reward model, $\cos(\cdot, \cdot)$ is the cosine similarity, and $m$ is a margin hyperparameter. This loss function pushes the representations of empathetic (question, answer) pairs closer together while pushing non-empathetic pairs apart.

During RL, this trained reward model provides a quality score for each generated answer $a_i$. This score is the cosine similarity between the embeddings of the question and the answer. A predefined threshold, $T$, is used to convert this score into a binary reward:
\begin{equation}
r_{\text{answer}} = 
\begin{cases} 
1, & \text{if } \cos(E(q), E(a_i)) > T \\
0, & \text{otherwise}
\end{cases}
\end{equation}
This mechanism directly rewards the generation of answers that are assessed to be genuinely empathetic and relevant, providing a robust signal for optimizing the therapeutic quality of the final response.

\section{Experiments}

\subsection{Experimental Setup}

\subsubsection{Corpora}

For evaluation, we conduct a comprehensive assessment of Empathy-R1 using two test sets: (1) 2,000 samples randomly selected from Empathy-QA (excluding training data) and (2) 2,000 samples from the publicly available PsyQA dataset. These test sets cover major psychological counseling topics, such as depression (35\%), anxiety (28\%), and interpersonal relationships (22\%), ensuring broad topic coverage for robust evaluation (see Appendix for details). Recognizing the open-ended nature of mental health counseling, where multiple responses can be equally valid and empathetic, we depart from the single golden answer paradigm. Instead, we adopt a more robust multi-reference evaluation strategy. For each question in both the Empathy-QA and PsyQA test sets, we utilize all available human-written responses as a reference set. To compute NLP-based metrics (e.g., BLEU, ROUGE), a model's generated response is scored against each individual reference in the set. The final score for the question is then calculated as the \textbf{arithmetic mean} of these scores. We posit that this methodology offers a fairer and more comprehensive assessment by acknowledging the diversity of appropriate therapeutic responses, rather than optimizing for a single, arbitrary ground truth.

\subsubsection{Training Configuration}

We conduct all experiments on a single node with 8$\times$NVIDIA A800 GPUs (80GB each). Our implementation uses Python 3, PyTorch, and the HuggingFace Transformers library. 
We adopt Qwen3-8B~\cite{qwen3} as our backbone model and disable its built-in thinking module throughout training to promote structured empathetic reasoning.
The training process consists of two stages. 
First, we perform Supervised Fine-Tuning (SFT) on the model with the LLaMA-Factory framework. 
Subsequently, we apply full-parameter Reinforcement Learning (RL) using the Verl framework~\cite{sheng2024hybridflow}. 
The RL training is distributed across 6 GPUs, with one additional GPU dedicated to reward model inference. 
We train for 1,000 steps using the GRPO algorithm and a custom reward. 
The learning rate for the actor is set to $1\mathrm{e}{-6}$, with micro-batches of size 1 per GPU and a global mini-batches of size 6.
Each input prompt generates $G=4$ candidate responses for reward comparison. 
Maximum input and output lengths are 1,024 tokens, and the rollout temperature is 1.0. To ensure reproducibility, a fixed random seed of 42 was used for all experiments.

\subsubsection{Baseline Models}
To ensure a comprehensive evaluation of Empathy-R1, we selected a diverse set of baselines: (1) the base model \textit{Qwen3-8B} in two configurations—\textit{Qwen3-8B\textsuperscript{T}} (with its native thinking module) and \textit{Qwen3-8B\textsuperscript{NT}} (without)—to isolate the impact of our framework; (2) prior state-of-the-art mental health models, \textit{CBT-LLM}~\cite{na2024cbt} and \textit{EmoLLM}~\cite{yang2024emollm}, as specialized baselines; and (3) a top-performing general-purpose model, \textit{Deepseek-R1}, known for its strong performance in Chinese, to compare against a powerful yet non-specialized alternative.

\subsubsection{Automatic NLP Metrics}

We adopt several standard metrics widely used in dialogue generation research, including word overlap-based metrics \textbf{BLEU-1} (\textbf{B-1})~\cite{papineni2002bleu}, \textbf{ROUGE-L} (\textbf{R-L})~\cite{lin2004rouge}, and \textbf{METEOR}~\cite{banerjee2005meteor}; and diversity metrics \textbf{Distinct-1} (\textbf{D-1})~\cite{li2015diversity}. To intuitively reflect the overall model performance, we also report a normalized average score (\textbf{NAvg}) across these metrics.

\subsubsection{Human Evaluation Design}

To capture the subjective quality of responses, we conducted a rigorous human preference evaluation on a 100-sample subset. We recruited 20 annotators from the general public (ages 15-60) to align with the intended audience. For each sample, annotators performed a \textbf{relative ranking} of all models' anonymized responses based on a \textbf{holistic judgment} of their overall quality. The presentation order was randomized to mitigate position bias. To ensure a consistent evaluative framework, this holistic judgment was guided by four key aspects: (1) \textbf{Fluency} (natural and coherent language, penalizing verbosity); (2) \textbf{Identification} (accurately recognizing the user's core dilemma and emotions); (3) \textbf{Comforting} (genuine, non-formulaic support creating a safe atmosphere); and (4) \textbf{Suggestion} (practical and constructive advice). This focus on a single holistic ranking, rather than scoring individual dimensions, was a deliberate methodological choice. Relative ranking is a more robust and reliable method for capturing overall user preference and has been shown to yield higher consistency than absolute scoring on nuanced tasks~\cite{novikova-etal-2017-need}. From these rankings, we report \textbf{Win-Rate@K} (the proportion of outputs ranked in the top-$K$ positions) and \textbf{Mean-Rank} (the average ranking position of each model).

\begin{table}[ht]
\centering
\caption{
Model performance on the Empathy-QA and PsyQA test sets. 
\textbf{Automatic NLP Metrics}: \textbf{B-1} = BLEU-1, \textbf{D-1} = Distinct-1, 
\textbf{R-L} = ROUGE-L, \textbf{MET.} = METEOR, \textbf{NAvg} = Normalized Average Score. 
\textbf{Human Preference Evaluation}: \textbf{Win@1} and \textbf{Win@2} represent the percentage of times the model was ranked 1st or in the top 2, respectively; 
\textbf{MR} denotes the Mean Rank (lower is better). 
Qwen3-8B$^{\mathrm{NT}}$ refers to the model without the "thinking" mechanism, while 
Qwen3-8B$^{\mathrm{T}}$ includes it.
}
\resizebox{\textwidth}{!}{%
\begin{tabular}{ll ccccc ccc}
\hline
\textbf{Test Set} & \textbf{Models} 
& \multicolumn{5}{c}{\textbf{Automatic NLP Metrics}} 
& \multicolumn{3}{c}{\textbf{Human Preference Evaluation}} \\ 
\cline{3-7} \cline{8-10}
 & 
 & \textbf{B-1↑} & \textbf{D-1↑} & \textbf{R-L↑} & \textbf{MET.↑} & \textbf{NAvg↑}
 & \textbf{Win@1↑} & \textbf{Win@2↑} & \textbf{MR↓} \\ 
\hline
\multirow[c]{7}{*}{Empathy-QA}  
           & \textbf{Empathy-R1 (Ours)} & \textbf{0.314} & 0.375 & 0.045 & \textbf{0.314} & 0.262 & \textbf{44.30} & \textbf{56.30} & \textbf{2.600} \\
           & Qwen3-8B$^{\mathrm{NT}}$   & 0.253 & 0.318 & 0.117 & 0.298 & 0.246 & 16.30 & 39.40 & 3.114 \\
           & Qwen3-8B$^{\mathrm{T}}$    & 0.250 & 0.315 & 0.111 & 0.299 & 0.244 & 11.40 & 34.00 & 3.257 \\
           & CBT-LLM                   & 0.251 & 0.449 & 0.013 & 0.270 & 0.246 & 10.90 & 30.60 & 3.411 \\
           & EmoLLM                    & 0.004 & \textbf{0.785} & 0.003 & 0.118 & 0.228 & 2.60  & 7.20  & 5.457 \\
           & Deepseek-R1               & 0.232 & 0.457 & \textbf{0.126} & 0.249 & \textbf{0.266} & 14.60 & 32.60 & 3.160 \\
\hline
\multirow[c]{7}{*}{PsyQA}     
           & \textbf{Empathy-R1 (Ours)} & \textbf{0.261} & 0.389 & 0.046 & \textbf{0.275} & \textbf{0.243} & \textbf{37.50} & \textbf{55.60} & \textbf{2.616} \\
           & Qwen3-8B$^{\mathrm{NT}}$   & 0.161 & 0.307 & \textbf{0.084} & 0.250 & 0.201 & 10.30 & 37.60 & 3.333 \\
           & Qwen3-8B$^{\mathrm{T}}$    & 0.161 & 0.307 & \textbf{0.084} & 0.249 & 0.200 & 9.50  & 25.00 & 3.296 \\
           & CBT-LLM                   & 0.257 & 0.451 & 0.009 & 0.241 & 0.240 & 12.30 & 29.20 & 3.298 \\
           & EmoLLM                    & 0.022 & \textbf{0.781} & 0.003 & 0.140 & 0.237 & 2.30  & 3.20  & 5.614 \\
           & Deepseek-R1               & 0.124 & 0.383 & 0.077 & 0.193 & 0.194 & 28.00 & 49.40 & 2.837 \\
\hline
\end{tabular}
}
\label{tab:evaluation}
\end{table}

\subsection{Results and Analysis}

\subsubsection{Automatic Evaluation Results}
We evaluated our model on the \texttt{Empathy-QA} and \texttt{PsyQA} test sets using standard metrics (\texttt{BLEU-1}, \texttt{ROUGE-L}, \texttt{METEOR}, \texttt{Distinct-1}) and a normalized average score (\texttt{NAvg}).

As shown in Table~\ref{tab:evaluation}, \textbf{Empathy-R1} excels in metrics measuring precision and semantic alignment. On the \texttt{Empathy-QA} test set, it achieves the best \texttt{BLEU-1} (0.314) and \texttt{METEOR} (0.314) scores, indicating strong lexical and deep semantic alignment with ground-truth answers. This trend holds on the \texttt{PsyQA} dataset, where \textbf{Empathy-R1} again leads in \texttt{BLEU-1} (0.261) and \texttt{METEOR} (0.275), demonstrating cross-domain robustness.

In terms of diversity, models like \texttt{EmoLLM} achieve an extremely high \texttt{Distinct-1} score (0.785 on \texttt{Empathy-QA}) but at a significant cost to relevance (0.004 \texttt{BLEU-1}). This pattern is attributed to \texttt{EmoLLM}'s design for short, brief conversations, leading to artificially high diversity scores that lack substance for the Long Counseling Texts (LCTs) central to our study. This highlights the limitation of relying on \texttt{Distinct-1} in isolation for evaluating long-form therapeutic dialogue.

This principled approach to generation provides context for the other metrics. \textbf{Empathy-R1} maintains a deliberate balance between lexical diversity and contextual relevance; its competitive \texttt{Distinct-1} scores (0.375 on \texttt{Empathy-QA}) are not inflated by trivial brevity. This is a direct result of our framework's design: the SFT phase instills the CoE scaffold, while GRPO encourages exploring diverse, high-quality expressions \textit{within} that structure. This prevents the model from collapsing into repetitive patterns while ensuring its outputs remain therapeutically grounded.

This balance, in turn, explains the \texttt{ROUGE-L} results. While models like \texttt{Deepseek-R1} show a notably higher \texttt{ROUGE-L} score, this metric's focus on the longest common subsequence can disproportionately reward models that simply reproduce lengthy phrases from the reference set. In contrast, \textbf{Empathy-R1}'s strong performance in \texttt{BLEU-1} and \texttt{METEOR}, coupled with its competitive \texttt{Distinct-1} score, suggests it generates responses that are lexically and semantically aligned with expert references while utilizing novel and diverse phrasing. This advanced capability—paraphrasing and restructuring ideas rather than merely copying them—naturally results in a lower \texttt{ROUGE-L} score, which we argue is a sign of more sophisticated generation rather than a deficiency.

Overall, the automatic evaluation results validate the effectiveness of our proposed \textbf{Empathy-R1} model in navigating the complex trade-off between structural fidelity and generative diversity. This quality, not fully captured by n-gram-based metrics, sets the stage for a more definitive human evaluation.

\begin{table}[!t]
\centering
\caption{Ablation Study of Training Strategy Components. The results demonstrate the synergistic effect of combining Supervised Fine-Tuning (SFT) and Group Relative Policy Optimization (GRPO). The best performance in each column is \textbf{bolded}.}
\begin{tabular}{lccccc}
\hline
\textbf{Method} & \textbf{B-1↑} & \textbf{D-1↑} & \textbf{R-L↑} & \textbf{MET.↑} & \textbf{NAvg↑} \\
\hline
Base Model      & 0.253 & 0.318 & \textbf{0.117} & 0.298 & 0.246 \\
+ SFT only      & 0.249 & 0.307 & 0.092 & 0.278 & 0.232 \\
+ GRPO only     & 0.292 & 0.354 & 0.066 & 0.309 & 0.255 \\
\textbf{Empathy-R1} & \textbf{0.314} & \textbf{0.375} & 0.045 & \textbf{0.314} & \textbf{0.262} \\ 
\hline
\end{tabular}
\label{tab:ablation_strategy}
\end{table}

\subsubsection{Human Preference Evaluation Results}

To address the limitations of automatic metrics, we conducted a rigorous human preference evaluation. The results (rightmost columns of Table~\ref{tab:evaluation}) show a substantial preference from human annotators for our \textbf{Empathy-R1} model over the state-of-the-art baselines.

A key indicator of this superiority is the \textbf{Win@1} rate, which signifies the percentage of times a model's response was ranked as the single best. On the \texttt{Empathy-QA} dataset, \textbf{Empathy-R1} achieves a Win@1 score of 44.30\%, a rate significantly higher than its closest competitor, Qwen3-8B$^{\mathrm{NT}}$ (16.30\%). This strong performance is also observed on the \texttt{PsyQA} dataset, where \textbf{Empathy-R1} again secures the highest Win@1 rate (37.50\%), outperforming the next-best model, Deepseek-R1 (28.00\%). This indicates our model's consistent ability to generate responses perceived as the top choice by human evaluators.

Furthermore, the \textbf{Mean Rank (MR)} provides a holistic view of a model's consistent performance. \textbf{Empathy-R1} secured the best (lowest) MR on both \texttt{Empathy-QA} (2.600) and \texttt{PsyQA} (2.616), indicating that its responses were consistently ranked higher than those from other models across the entire evaluation set. It is particularly noteworthy to contrast our model with competitors like \texttt{CBT-LLM}. While \texttt{CBT-LLM} shows a competitive Win@2 rate, its much lower Win@1 score suggests it frequently produces responses that are acceptable but not exceptional. In stark contrast, \textbf{Empathy-R1}'s profile—characterized by a high Win@1 and a strong Win@2—suggests it is more effectively optimized to produce standout, first-choice answers rather than merely "good enough" alternatives.

In summary, the consistent human preference for \textbf{Empathy-R1} validates that the advancements driven by our framework generate qualitatively superior responses aligned with genuine human expectations.

\begin{table}[!t]
\centering
\caption{Ablation Study of the Chain-of-Empathy (CoE) Reasoning. The table compares the performance of our model with the CoE reasoning paradigm (w CoE) against a variant trained without it (w/o CoE).}
\begin{tabular}{lccccc}
\hline
\textbf{Method} & \textbf{B-1↑} & \textbf{D-1↑} & \textbf{R-L↑} & \textbf{MET.↑} & \textbf{NAvg↑} \\
\hline
w CoE & \textbf{0.314} & \textbf{0.375} & 0.045 & \textbf{0.314} & \textbf{0.262} \\
w/o CoE & 0.296 & 0.345 & \textbf{0.054} & 0.309 & 0.251 \\
\hline
\end{tabular}
\label{tab:coe_compare}
\end{table}

\subsection{Ablation Studies}

\subsubsection{Analysis of Training Strategy Components}

To analyze the effectiveness of our training pipeline, we conducted an ablation study with the pre-trained Qwen3-8B model serving as the base model for all subsequent experiments. As shown in Table~\ref{tab:ablation_strategy}, applying only Supervised Fine-Tuning (SFT) resulted in a performance drop across all metrics. This is because forcing the model to strictly adhere to the Chain-of-Empathy format via SFT alone makes its output rigid and compromises overall response quality. Furthermore, while using only Group Relative Policy Optimization (GRPO) shows improvement, it is not as effective as the combined SFT+GRPO approach. Through observation of the outputs, we found this is because the pure GRPO training process is more unstable without the SFT phase to first establish the structured reasoning framework. This highlights a clear synergy: SFT provides the necessary structural foundation, which GRPO then refines to achieve optimal performance.

\subsubsection{Effectiveness of Chain-of-Empathy Reasoning}

To validate the contribution of our Chain-of-Empathy (CoE) reasoning paradigm, we conducted a targeted ablation study. We compared our full Empathy-R1 model (w CoE) against an identical model trained without the CoE-structured thinking process (w/o CoE). For the `w/o CoE` variant, we removed the four-layered analysis prompt (L1-L4), allowing the model to generate responses freely based only on the general instruction to act as a counselor. The results, detailed in Table~\ref{tab:coe_compare}, highlight the pivotal role of CoE in enhancing response quality.

The model equipped with CoE (`w CoE`) significantly outperforms its counterpart (`w/o CoE`) on key metrics measuring semantic relevance and lexical alignment, including a notable increase in both BLEU-1 (0.314 vs. 0.296) and METEOR (0.314 vs. 0.309). This indicates that by guiding the model through a structured analysis of emotions, causes, and intents, CoE produces responses that are more closely aligned with professional human-written answers. Furthermore, the higher Distinct-1 score (0.375 vs. 0.345) suggests that this structured reasoning process encourages more diverse and less repetitive generation.

Interestingly, the model without CoE achieves a slightly higher ROUGE-L score. We interpret this not as a deficiency of CoE, but rather as evidence of its advanced capability. A higher ROUGE-L can reward models for simply reproducing long, contiguous phrases from the reference texts. In contrast, the CoE framework compels the model to first deconstruct the user's problem and then synthesize a novel, structured response. This act of genuine reasoning and re-phrasing naturally results in a lower longest-common-subsequence score but represents a more sophisticated and therapeutically valuable form of generation. In essence, CoE trades superficial string matching for deeper, more meaningful empathetic engagement, confirming its effectiveness as the core component of our framework.

\section{Related Work}

\subsection{LLMs for Mental Health Support}

The application of Large Language Models (LLMs) to mental health support has seen rapid growth, evolving from rule-based chatbots to sophisticated, fine-tuned models \cite{dhingra2023mind, zheng2023building}. A prevalent methodology involves fine-tuning on domain-specific corpora, and prominent examples like MentaLLaMA and CBT-LLM have demonstrated the potential of this approach \cite{yang2024mentallama, xu2024mental, na2024cbt}. However, a significant gap remains for Long Counseling Texts (LCTs). This reveals a critical flaw in the current paradigm: fine-tuning enhances a model's ability to match patterns and retrieve therapeutic knowledge, but it does not inherently endow it with a structured, human-like reasoning process \cite{lee2023chain, jung2025ve}. Consequently, translating cognitive empathy into a coherent and therapeutically beneficial response remains a pressing, unsolved challenge, setting the stage for a new class of structured reasoning paradigms.

\subsection{Chain-of-Thought and Reasoning in LLMs}

The introduction of Chain-of-Thought (CoT) prompting was a significant breakthrough for eliciting multi-step reasoning, shifting research focus from solely scaling models to explicitly scaffolding their internal deliberation process \cite{wei2022chain}. This foundational concept has inspired more sophisticated frameworks like Tree-of-Thoughts (ToT) and Graph-of-Thoughts (GoT), which treat problem-solving as a structured search process \cite{yao2023tree, besta2024graph}. These methods establish a powerful principle: decomposing a complex task into a structured workflow is key to enhancing the reliability and transparency of LLM-generated outputs. However, while these techniques have proven highly effective in formal domains like mathematics and logic, their application to domains characterized by ambiguity, emotional nuance, and subjective human experience remains a largely unexplored frontier.

\section{Conclusion}

This paper introduced \textbf{Empathy-R1}, a novel framework that endows LLMs with deep, psychologically-grounded reasoning for mental health support via its \textbf{Chain-of-Empathy (CoE)} paradigm. Extensive experiments demonstrated the superiority of Empathy-R1. Most notably, in rigorous human preference evaluations, our model achieved a commanding \textbf{Win@1 rate of 44.30\%}, a figure nearly four times higher than its closest competitor. This result unequivocally validates that Empathy-R1 generates responses perceived as significantly more helpful, relevant, and genuinely empathetic. In conclusion, Empathy-R1 marks a significant advancement toward developing responsible, scalable, and genuinely beneficial AI for mental health support, paving the way for a new generation of AI systems capable of providing deeper and more meaningful care.

\bibliographystyle{abbrvnat} 
\bibliography{main}

\end{document}